\documentclass[10pt,conference]{IEEEtran}
\IEEEoverridecommandlockouts
\usepackage{cite}

\usepackage{amsmath,amssymb,amsfonts,mathtools, bm}
\usepackage{amsthm}
\usepackage{algorithm}
\usepackage{algorithmic}
\usepackage{graphicx}
\usepackage{textcomp}
\usepackage{xcolor,float}
\usepackage{tabularx}
\usepackage{textcomp}
\usepackage{diagbox}
\usepackage{svg}
\usepackage{booktabs}
\usepackage{subfigure}
\usepackage{hyperref}
\newcounter{MYtempeqncnt}

\allowdisplaybreaks
\renewcommand{\baselinestretch}{1}

\begin{document}

\title{Imperfect Digital Twin Assisted Low Cost Reinforcement Training for Multi-UAV Networks}
\author{
\IEEEauthorblockN{
Xiucheng Wang\IEEEauthorrefmark{1},
Nan Cheng\IEEEauthorrefmark{1},
Longfei Ma\IEEEauthorrefmark{1}, 
Zhisheng Yin\IEEEauthorrefmark{2},
Tom. Luan\IEEEauthorrefmark{3},
Ning Lu\IEEEauthorrefmark{4}\\
}
\IEEEauthorblockA{
\IEEEauthorrefmark{1}School of Telecommunications Engineering,
Xidian University, Xi'an, China\\
\IEEEauthorrefmark{2}School of Cyber Engineering,
Xidian University, Xi'an, China\\
\IEEEauthorrefmark{3}School of Cyber Science and Engineering, Xi’an Jiaotong University, China\\
\IEEEauthorrefmark{4}Department of Electrical and Computer Engineering, Queen’s University, Ontario, Canada\\
Email: \{xcwang\_1, lfma\}@stu.xidian.edu.cn, dr.nan.cheng@ieee.org, zsyin@xidian.edu.cn,\\ tom.luan@xjtu.edu.cn, ning.lu@queensu.ca}}
    \maketitle

\IEEEdisplaynontitleabstractindextext

\IEEEpeerreviewmaketitle

\begin{abstract}
Deep Reinforcement Learning (DRL) is widely used to optimize the performance of multi-UAV networks. However, the training of DRL relies on the frequent interactions between the UAVs and the environment, which consumes lots of energy due to the flying and communication of UAVs in practical experiments. Inspired by the growing digital twin (DT) technology, which can simulate the performance of algorithms in the digital space constructed by coping features of the physical space, the DT is introduced to reduce the costs of practical training, e.g., energy and hardware purchases. Different from previous DT-assisted works with an assumption of perfect reflecting real physics by virtual digital, we consider an imperfect DT model with deviations for assisting the training of multi-UAV networks. Particularly, to trade off the training cost, DT construction cost, and the impact of deviations of DT on training, the real and virtually generated UAV mixing deployment method is proposed, and two cascade neural networks (NN) are used to jointly optimize the number of virtually generated UAVs, the DT construction cost, and the performance of multi-UAV networks. These two NNs are trained by unsupervised learning and reinforcement learning, which are both low-cost label-free training methods. Simulation results show the training cost can be significantly decreased while guaranteeing the training performance, which implies that an efficient decision can be made with imperfect DTs in multi-UAV networks.
\end{abstract}


\section{Introduction}
Unmanned aerial vehicles (UAVs) are widely used to enhance the performance of networks by serving as access points \cite{8660516}. Since UAVs are movable, it is important to design the flying trajectory, thus optimizing the average network performance in temporal \cite{8434285}. Traditionally, temporal optimization can be solved by dynamic programming (DP). However, as the number of UAVs in the network increases, the dimension of state space in DP grows dramatically. Therefore, it is time-consuming to search the total state space to find the optimal trajectory using DP when there are multiple UAVs in the network. Fortunately, the deep neural network (DNN) based reinforcement learning (RL) method can effectively optimize UAVs' trajectory by only computing several times matrix multiplication. Thus, nowadays, lots of work focus on using deep RL (DRL) to solve the optimization of UAVs in the wireless network \cite{cheng2019space,wang2022joint,9127423}.

However, training a DRL agent for the UAVs network can be costive due to the high costs of hardware and data transmission. Obviously, to train a DRL agent for multiple UAVs-assisted wireless networks, it is necessary to deploy a certain number of UAVs in the network and interact with the environment. Generally, training DRL relays on lots of data that is obtained through UAVs' interaction with the environment. Therefore, during the training procedure, multiple UAVs need to experiment with a large number of different flight trajectories for obtaining a large amount of training data in the interaction with the environment, which leads to a huge energy consumption of all UAVs. Particularly, a centralized controller is needed to collect information about the environment and UAVs to generate the training input and reward of the DRL agent. Training a DRL agent usually needs a large amount of data, which leads to a huge consumption of both time and bandwidth for transmitting so much data to the controller. 
\begin{figure*}[h]
  \centering
  \includegraphics[width=1.5\columnwidth]{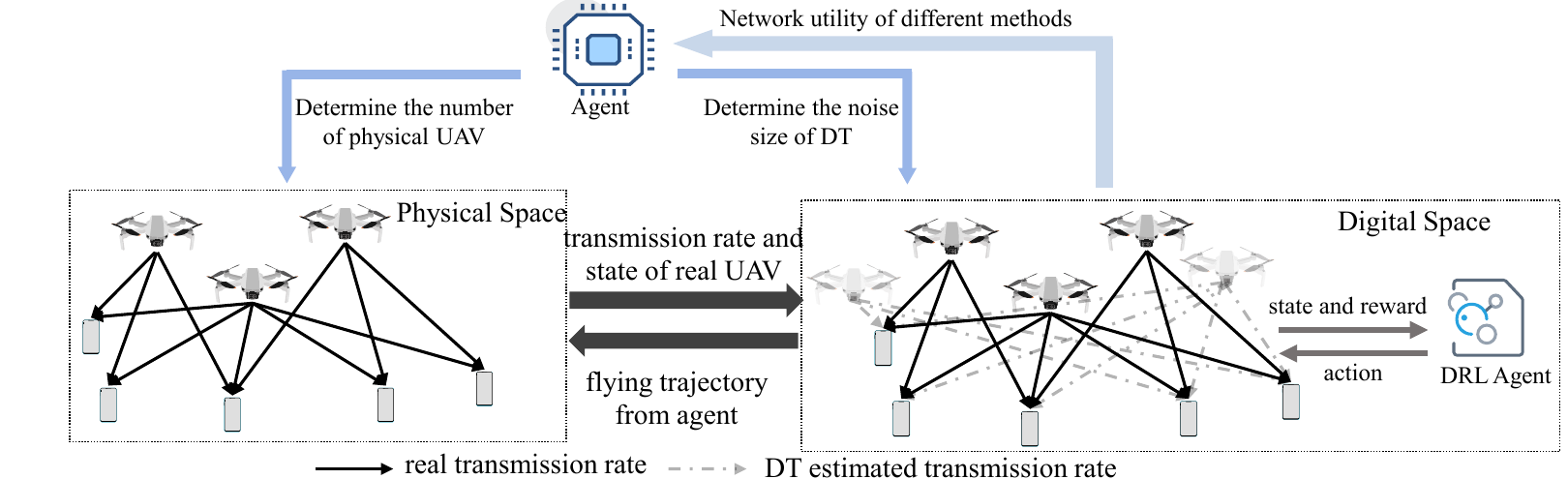}
   \vspace{-5pt}
  \centering \caption{The illustration of the system model for five UAVs network, where three UAVs are deployed in the physical space and the rest two UAVs are generated virtually in the digital space.}
  \label{system}
   \vspace{-5pt}
\end{figure*}
It will be low-cost if the training of UAV networks does not rely entirely on multiple UAVs interacting with the physical environment but on a certain simulation environment. Fortunately, the rapidly evolving digital twin (DT) technology provides a method to reproduce the information in physical space using digital technology simulation \cite{wu2021digital,9263396}. With DT, UAVs can analyze what will happen in the physical space by interacting with the digital environment. Then RL agent can use the data and reward generated in the DT to learn how to optimize the trajectory of UAVs. In this case, both the cost of buying UAV hardware and the energy consumption required to train the UAVs can be saved. Moreover, the RL agent can also be trained in the digital space, thus the training data no longer needs to be transmitted from UAVs to the controller, and the data transmission cost can also be saved.

In previous works, the DT is usually assumed as perfect in wireless networks, which can perfectly build a digital copy of the physical space, and there is no deviation between DT and the real world \cite{9860495,nguyen2021digital}. However, limited by the technology the copying error cannot be avoided between digital space and physical space, which indicated the DT cannot perfectly reflect all features of entities in physical space or represent that with errors. Therefore, it may cannot correctly predict what will happen when an action is taken in a specific state. Remarkably, the deviation between DT and real work will cause the transfer loss of the DRL agent, which means if the DRL agent directly uses the knowledge learned from the DT, the performance of the UAV network will be decreased. Therefore, to reduce both the training cost of RL and the construction cost of DT, it is necessary to design both economic and efficient reinforcement training methods assisted by the imperfect DT. 

In this paper, to trade off the training performance caused by DT transfer loss and cost of DT contraction, UAV hardware cost, and data synchronization between physical space and digital space, a imperfect DT-assisted RL training method is proposed. Specifically, a multi-UAV network is considered, however, in the training procedure only a part of the UAVs are deployed in the physical space, while the rest of the UAVs are virtually generated in the digital space by DT. Therefore, the low cost of training in digital space and the high accuracy of training in the physical space can be both exploited. 

Remarkably, the construction cost of DT is also considered, therefore it is also essential to determine how precise the DT needs to be since higher accuracy leads to both higher training performance and construction cost. To deal with the above challenges, an MLP trained by unsupervised learning is used to determine the number of physical UAVs and the construction precision of DT, which are regarded as hyper-parameters to train the DRL for optimizing the UAV trajectory.

The main contributions of this paper are as follows.
\begin{enumerate}
    \item A imperfect DT-assisted low-cost reinforcement training method for the UAV network is  proposed, where the cost of DT contraction, UAV hardware cost, and data synchronization between physical space and digital space are all considered as training costs. 
    \item To exploit the both benefits of the low cost in digital space and the high accuracy in physical space in DT, a real and virtually generated UAVs mixed deployment is proposed.
    \item The joint optimization of DRL training cost and performance are solved by two cascade NNs, which are used 
    to determine the number of virtual generated UAVs and the construction accuracy of DT, and to optimize the UAVs' flying trajectory, respectively. The two NNs are trained by unsupervised learning and reinforcement learning, which are label free.
\end{enumerate}

\section{System Model}
\subsection{Transmission Model}
In this paper, we consider there are $K$ users located in remote village or areas where the infrastructure are absent or destroyed. Due to their effectiveness and flexibility, $M$ UAVs are used as movable access points and provide communication services to users to maximize the average communication rate for any user.

Inspired by the success of orthogonal frequency division multiplexing (OFDM) in 4G/5G networks, in this paper OFDM is used to remove the interference among users and UAVs. According to \cite{mao2017joint}, the transmission between user $i$ and UAV $j$ can be calculated as
\begin{align}
    rate^{real}_{ij} = Blog_2\left(1+\frac{g_0\left (L_0/L_{ij}\right)^{\theta}p}{\sigma^2}\right),
\end{align}
where the $B$ is the bandwidth and $g_0$ is the path-loss constant, $L_0$ is the reference distance and $L_{ij}=\|loc^{user}_{i}-loc^{uav}_j\|$ is the distance between user $i$ and UAV $j$ and $loc^{user}_{i}$ is the location of user $i$ $loc^{uav}_{j}$ is the location of UAV $j$, $p$ is the transmission power and $\sigma^2$ is the noise power. 

Assuming all $K$ users are randomly distributed in a finite plane and the location of users will not be changed, and all UAVs are initially located in the same place, which can be regarded as the hangar of UAVs. Therefore, it is necessary to optimize the location of UAVs to increase users' transmission rate, and the flying trajectory to enhance the average time transmission rate for users. Similar to \cite{8434285}, the location of UAVs is optimized by changing the flying direction $v$ with a constant flying speed in each slot. In this paper we consider the scenario that the distance between any user and UAV is not very far, thus a user may establish a communication link with multiple UAVs. Since OFDM is used and the interference is neglected, the communication rate for a single user is the maximum rate between a specific user and all UAVs. 

\subsection{Imperfect Digital Twin Model}
To effectively optimize the flying trajectory imperfect DT is used, which has three main functions physical data reproduction, simulation generation, and decision control. Besides advantages, the construction of DT is also considered that the more powerful the DT is, the higher the construction cost is.

Firstly, DT can collect data from physical space and reproduce the features of physical entities in digital space by certain technical methods. Since features of the digital copy should always be consistent with the physical entities, it is necessary to frequently transmit entities information from the physical space to the DT, to ensure all features changes in the physical space can be reflected in the digital space in real-time \cite{9040431}. Due to the frequent information synchronization, the transmission cost cannot be simply neglected. Assuming the information transmission frequency of all physical entities is the same, thus the transmission cost is determined by the number of physical entities, more entities lead to higher transmission costs.

In addition to reproducing information in physical space, DT can also analyze the interaction of different actions of different entities with the environment in the digital simulation space, even if such actions and entities do not really happen in the physical space \cite{wu2021digital}. Therefore, if there are only $K$ UAVs deployed in the physical space, the DT can generate the extra $M-K$ UAVs to analyze the different trajectory influences when there are $M$ UAVs. However, limited by the technology and the complexity of predicting the future, the simulated result cannot be exactly the same as the real physical. In this paper, we consider the scenario that if a UAV is generated in digital space and does not have a physical entity corresponding to it, then there is a noise bias in the calculation of the transmission rate of that UAV to the user, where the power of noise is related to the construction cost of DT.

Generally, the DT is equipped with abundant computing resources, therefore the UAV trajectory controller is deployed in the digital space to reduce the decision latency. Moreover, since the digital space contains all the environmental information, including the information collected from the physical space and the data from the extra generated UAVs in the digital space, deploying the controller in the digital space eliminates the cost of transmitting data to the controller.

\section{Mixed UAVs Deployment Method and Problem Formulation}
In this section, the DRL is proposed to optimize the trajectory of UAVs, and to reduce the training cost the virtual and physical UAVs mixed deployment method is proposed, then the problem of jointly increasing the transmission rate for all users in average time and decreasing the training cost is formulated.
\subsection{Virtual and physical UAVs Mixed Deployment}
Traditionally, to train a DRL agent for optimizing the trajectory for a network with $M$ UAVs, it is necessary to deploy an equal number of UAVs in the physical space, and then try different flight trajectories and constantly interact with the environment to evaluate the action of DRL agent. Obviously, such a training method leads to high hardware costs in buying UAVs and the energy consumption for the agent needs to try many different trajectories to converge, which leads to high flying energy costs. Fortunately, the DT can evaluate the performance of flying trajectory in the digital space, even though there are no UAVs deployed in the physical space. However, since the DT is considered imperfect that the analyzing result may have differences from the real physical situation. In this paper, we consider the scenario in which the simulated transmission rate is different from the real transmission rate by adding noise. Thus the transmission rate between virtually generated UAVs $j$ and user $i$ is
\begin{align}
    rate^{dt}_{ij} = Blog_2\left(1+\frac{g_0\left (L_0/L_{ij}\right)^{\theta}p_i}{\sigma^2}\right)+\mathcal{N}(0,\delta),
\end{align}
where $\mathcal{N}(0,\delta)$ is a Gaussian random variable with 0 as the mean and $\delta$ as the variance, and the size of $\delta$ is related to the construction cost of DT, the smaller $\delta$ is the larger the construction cost of DT. To formulate the construction cost $\epsilon$, similar to \cite{khemka2014utility} the construction cost is considered as exponentially increase as the noise $\delta$ decrease
\begin{align}
    \epsilon = \alpha e^{\beta\delta},
\end{align}
where $\alpha$ and $\beta$ are constant factors.

To deal with the challenges of imperfection in DT and the hardware cost in training, the virtual and physical UAV mixed deployment method is proposed. As is shown in Fig.~\ref{system}, to train an agent for control $M$ UAVs, only $K$ UAVs are needed to deploy in the physical space, while others can be generated by the DT. There is no hardware cost for virtual UAVs generated by DT since it does not need to buy physical UAVs and consume the energy to try different flying trajectories, thus the training cost can be reduced. Meanwhile, the deployment of physical UAVs increases the DRL training performance by reducing the simulation noise. However, more physical UAVs also lead to higher data transmission costs between physical and digital space. Assuming the data synchronization cycle between physical and digital space is a constant number, and the training iterations for the DRL agent are the same in all episodes, thus each physical UAV has the same distance to fly and the same amount of data to transmit. As a consequence, the cost of deploying UAVs in physical space is linearly related to the number of UAVs.

\subsection{Problem Formulation}
Above all, there are two main challenges in the imperfect DT-assisted UAV network to train the DRL agent: optimizing the flying trajectory to increase the network performance and reducing the training cost. Since the mixed virtual and physical UAVs deployment is used to fully use the benefits of physical and digital space, the DT construction cost is also considered, there are three main optimization variables: UAVs flying direction in each slot $\mathbf{v}$, the number of UAVs deployed in the physical space $K$, and the DT construction cost determined by the noise power $\delta$. Therefore, the problem can be formulated as
\begin{subequations}
\begin{align}
    &\max_{\mathbf{v},\delta,K}\; \eta\sum_{t}\sum_{i=1}^{N}\max_{j\in\{1,2,\cdots,M\}} log_2\left(1+\frac{g_0\left (L_0/L^{t}_{ij}\right)^{\theta}p_i}{\sigma^2}\right) \notag\\
    &\qquad\; -\alpha e^{\beta\delta}-\zeta K,\label{obj}\\
    &s.t.\quad L^{t}_{ij} = \|\mathbf{loc}^{user}_{i}-\mathbf{loc}^{uav,t}_{j}\|,\label{c1}\\
    &\qquad\; \|\mathbf{v}^t\|=v,\label{c2}\\
    &\qquad\; \mathbf{v} = \mathcal{Q}(\mathbf{loc}^{user},\mathbf{loc}^{uav,t-1}\,|\,\delta),\label{c3}\\
    &\qquad\; \mathbf{loc}^{uav,t}_j = \mathbf{loc}^{uav,t-1}_j +\mathbf{v}^t,\label{c4}\\
    &\qquad\;  K\in \{0,1,\cdots,M\},\label{c5}\\
    &\qquad\; 0\le \delta,\label{c6}
\end{align}
\end{subequations}where the $\eta$, $\zeta$, $\alpha$ and $\beta$ are all constant factors. The objective function (\ref{obj}) is to jointly maximize the transmission rate for all users in average time and reduce the training cost including buying UAVs, and UAVs flying energy consumption and DT construction. Constraint (\ref{c1}) is the distance between user $i$ and UAV $j$, and the constraint (\ref{c2}) guarantees the size of flying speed is a constant number. In Constraint (\ref{c3}) the $\mathcal{Q}(\cdot|\cdot)$ is the NN used in the DRL agent, it inputs the location of users and UAVs to determine the flying direction of UAVs in slot $t$, however, the noise $\delta$ affects the parameters of the NN $\mathcal{Q}$ by influencing the training loss, which affects the performance of $\mathcal{Q}$. The constraint (\ref{c4}) is the updates of the location of UAVs, and the constraint (\ref{c5}) shows the number of physically deployed UAVs is up to $M$, while the minimum number of physical UAVs is $0$. The constraint (\ref{c6}) shows the deviation $\delta$ between physical space and digital space in DT.

\section{Low-Cost Imperfect DT Assisted DRL Training Method}
Focusing on the above problem, there are three optimization variables, i.e., UAVs flying direction in each slot $\mathbf{v}^t$, the number of physically deployed UAVs $K$, and the noise of DT simulation $\delta$. Obviously, only $\mathbf{v}^t$ needs to be changed in each time slot which can be optimized by the DRL agent, while $K$ and $\delta$ can be regarded as hyper-parameters when training the DRL agent to optimize the flying trajectory. However, the challenge is that $K$ and $\delta$ not only determine the cost of training energy consumption, hardware purchases, and DT construction but also affect the training performance of the DRL. Besides, the performance of the DRL will in turn affect choosing the optimal $K$ and $\delta$. To deal with this challenge, two NNs are used in series, the first NN $\mathcal{G}$ is trained by the unsupervised method to determine the $K$ and $\delta$ trade-off the training cost and trajectory optimization performance, then the next NN $\mathcal{Q}$ is trained by the DRL method to optimize the flying trajectory under the certain $K$ and $\delta$ output by the $\mathcal{G}$.

Different from the $\mathcal{G}$ that optimizes the timing serial variable $\mathbf{v}^t$, which can be trained by the DRL method, the $K$ and $\delta$ are both scalar variables and they are constant in time, therefore it is challenging to model the optimization of $K$ and $\delta$ as a Markov decision process (MDP). Moreover, it is time-consuming to obtain the optimal values of $K$ and $\delta$ as the label to train the $\mathcal{G}$ by supervised learning. Fortunately, the objective function (\ref{obj}) is differentiable, thus the parameters $\theta_{\mathcal{G}}$ updating derivative in NN $\mathcal{G}$ can be obtained through the chain rules of derivative by calculating the derivative of the objective function (\ref{obj}) concerning $K$ and $\delta$ first, which are outputs of $\mathcal{G}$. Thus, by denoting the objective function (\ref{obj}) as $\mathcal{H}(\mathbf{v},\delta, K)$ we can update the $\theta_{\mathcal{G}}$ by gradient descent as Equation (\ref{gra-g}), where the $lr_{\mathcal{G}}$ is the learning rate to update parameters in $\mathcal{G}$. The Equation (\ref{gra-g}) couples the NN $\mathcal{G}$ and $\mathcal{Q}$ by the gradient that the output $\mathbf{v}$ output by $\mathcal{Q}$ affects the parameters updating gradient of $\mathcal{G}$. The higher performance of trajectory optimization performance of $\mathcal{Q}$, the higher accuracy of the parameters gradient of $\mathcal{G}$, then the $\mathcal{G}$ can make a better trade-off between training cost and training performance of $\mathcal{Q}$. Meanwhile, the trade-off of the $\mathcal{Q}$ performance, in turn, affects the gradient calculation accuracy of $\mathcal{G}$. Such a training method enforces the $\mathcal{G}$ to learn the joint optimization of $\mathbf{v},\delta, K$ in the upper layer with low training challenge, since the optimization of $\mathbf{v}$ is determined by another NN $\mathcal{Q}$.
\begin{figure*}[!t]
\normalsize
\setcounter{MYtempeqncnt}{\value{equation}}
\begin{align*}
&\theta_{\mathcal{G}} = \theta_{\mathcal{G}}+lr_{\mathcal{G}}\nabla_{\theta_{\mathcal{Q}}}\mathcal{G}(\alpha,\beta|\mathcal{Q}))\nabla_{\delta,K}\mathcal{H}(\mathbf{v},\delta,K)|_{\mathbf{v}=\mathcal{Q}(\mathbf{loc}^{user},\mathbf{loc}^{uav}),\{\delta,K\}=\mathcal{G}(\alpha,\beta|\mathcal{Q})}, \tag{6}\label{gra-g}\\
&\theta_{Q} = \theta_{\mathcal{Q}}+lr_{\mathcal{Q}}(r_t + max_{\mathbf{v}_{t+1}}Q^{-}(s_{t+1},\mathbf{v}_{t+1})-Q(s_{t},\mathbf{v}_{t})\nabla_{\theta_{\mathcal{Q}}}\mathcal{Q}(s_{t},\mathbf{v}_{t})),\tag{7}\label{gra-q}
\end{align*}
\setcounter{equation}{\value{MYtempeqncnt}}
\hrulefill
\vspace*{4pt}
\end{figure*}

After training, the $\mathcal{G}$ is used to determine the $K$ and $\delta$, and the trajectory optimization NN $\mathcal{Q}$ is trained by the DRL method, since the flying direction $\mathbf{v}$ is a timing serial variable, under the hyper-parameters $K$ and $\delta$. The trajectory control of UAVs is a continuous decision process, and the optimal direction of each time slot is determined by the current network state. Therefore, the process can be modeled as a Markov decision process (MDP) and represented as a tuple $<S, A, p, r, \gamma>$, where the meaning of each element is
\begin{itemize}
\item $S$ indicates the state space, where $s_{t} \in S$ denotes the state at time $t$.
\item $A$ denotes the action space of the agent, where $a_{t} \in S$ denotes the action at time $t$.
\item $p$ is the state transfer probability distribution.
\item $r$ denotes the reward, i.e., the environmental feedback after an action is executed by the agent.
\item $\gamma$ is the discount factor, which indicates the importance given to future rewards.
\end{itemize}
Based on this, a DQN-based algorithm is proposed to solve the multi-UAV trajectory decision problem. We assume that the DRL agent located in digital space controls the flight direction of all UAVs in the network. At the beginning of each time slot, the agent gives the flight direction of each UAV according to the network state and then improves the strategy based on the reward from the network. Among them, the network state is the position of users and UAVs. The reward is the sum rate of users at the current state, denoted as 
\begin{align}
    r_{t} = &\sum_{i}^{N}\max_{j\in\{1,2,\cdots,M\}} log_2\left(1+\frac{g_0\left (L_0/L^{t}_{ij}\right)^{\theta}p_i}{\sigma^2}\right)\notag\\
    &+\mathcal{N}(0,K\delta), 
\end{align}

To reduce the action space, we discretize the flight direction $\mathbf{v}$ of each UAV as $\{00,01,10,11\}$, which indicated the flying direction as $\{0^\circ,90^\circ,180^\circ,270^\circ\}$. In addition, to improve the performance, we bring in a target Q-network $\mathcal{Q}^-$ to help train the $\mathcal{Q}^-$, and the parameters $\theta_{\mathcal{Q}}$ are updated according to Equation (\ref{gra-q}), where the $lr_{\mathcal{Q}}$ is the learning rate for $\mathcal{Q}$.

\section{Simulation Results}
In this section, we evaluate the proposed method by setting $N=100$, $M=4$, $p=100$ mW, $\sigma^2=-174$ dBm/Hz. In addition, all UAVs have a height of 5 m and a speed of 8 m/s. The simulation results of the proposed algorithm are presented in two aspects. Firstly, we evaluate the convergence performance of different schemes. Secondly, we compare the network utility of the proposed scheme with other benchmark schemes by considering the sum rate and cost. To evaluate the performance of the proposed method, we use the following algorithm for comparison: (1) training the DRL using the UAVs fully deployed in the physical space; (2) training the DRL with a fixed number of UAVs deployed in the physical space while the rest is generated in digital space; (3) the proposed joint optimization of the number of UAVs deployed in the physical space, DT noise and the UAVs trajectory designing.

\subsection{Convergence Performance}

\begin{figure}[ht]
  \centering
  \includegraphics[width=0.78\columnwidth]{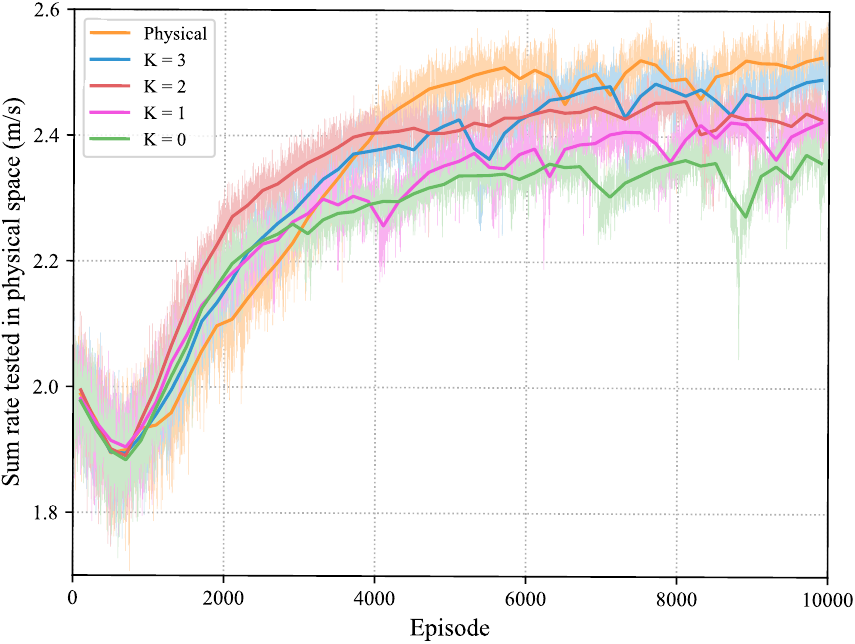}
   \vspace{-5pt}
  \centering \caption{Convergence performance of physical UAV and different number of DTs.}
  \label{fig1}
   \vspace{-5pt}
\end{figure}

\begin{figure}[ht]
  \centering
  \includegraphics[width=0.78\columnwidth]{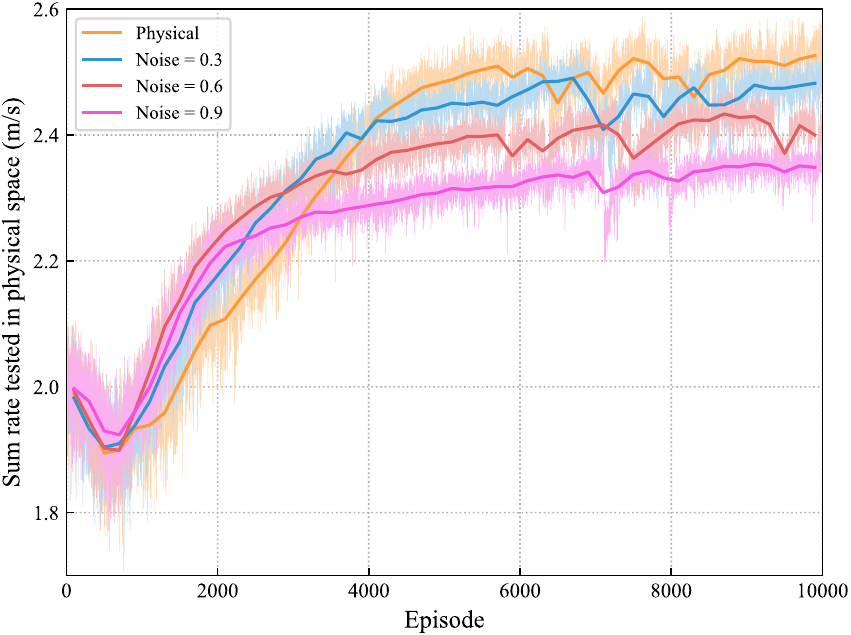}
   \vspace{-5pt}
  \centering \caption{Convergence performance of physical UAV and DT with a different noise.}
  \label{fig2}
   \vspace{-5pt}
\end{figure}
To demonstrate the performance in a real deployment scenario, we test the sum rate in the physical space.
Fig. \ref{fig1} shows the convergence performance of physical UAVs and different numbers of DTs, where the darker line is the moving average result of the last 200 episodes. All DTs in the figure has a noise size of 0.8. It can be observed that the convergence performance using DTs for training is inferior to that of the physical UAV. this is because the noise of the DTs interferes with the learning of the intelligence and makes it difficult for the intelligence to find the optimal behavioral strategy. In addition, it can be observed that the convergence performance decreases as the number of DTs increases. With the same noise of each DT, increasing the number of DTs makes the system noisier and thus the performance deteriorates. 

Figure \ref{fig2} shows the convergence performance of physical UAV and DT with different noise, where four DTs are used except for the case of the physical UAV. It can also be seen that the convergence performance of DTs is not as good as that of the physical UAV, and the system performance deteriorates with increasing noise.

\subsection{Network Utility Evaluation}

\begin{figure}[ht]
  \centering
  \includegraphics[width=0.78\columnwidth]{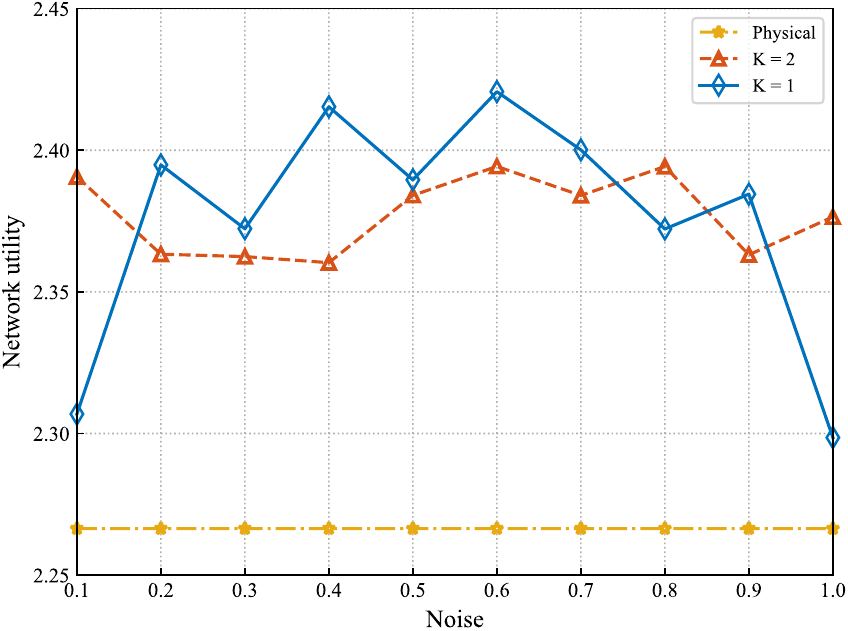}
   \vspace{-5pt}
  \centering \caption{Network utility comparison of physical UAV and different DTs.}
  \label{fig3}
   \vspace{-5pt}
\end{figure}

Fig. \ref{fig3} shows the network utility comparison of physical UAVs and different DTs. It can be observed that training with DT outperforms the physical UAV. Although the physical UAV is not affected by the noise, its high cost reduces the network utility. Moreover, it can be observed that the noise size of DT influences the network utility significantly. This is because although constructing a more accurate DT can reduce noise interference, it also increases the system's cost. Therefore, it is necessary to find the optimal DT based on the network characteristics.

\begin{figure}[ht]
  \centering
  \includegraphics[width=0.78\columnwidth]{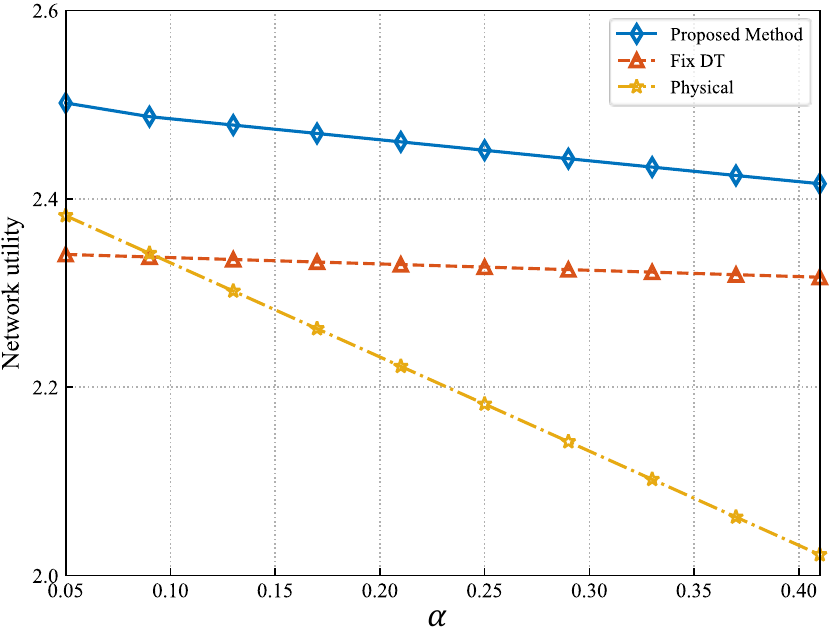}
   \vspace{-5pt}
  \centering \caption{Network utility v.s. $\alpha$.}
  \label{fig4}
   \vspace{-5pt}
\end{figure}

Fig. \ref{fig4} shows the network utility with respect to the weight on the cost, i.e., $\alpha$. The fixed DT in the figure uses 4 DTs with a noise size of 0.9. It can be observed that the network utility of each method decreases with increasing cost weights. The network utility of physical UAVs decreases the fastest as it has the largest cost. In the case of low-cost weights, the network utility of fixed DT is inferior to that of physical UAVs because of its weaker performance. In the case of high-cost weights, the network utility of fixed DT is better than that of physical UAVs because of its lower cost. The RL-based DT approach is able to balance the system performance and cost well, and thus achieves the optimal network utility.

\begin{figure}[ht]
  \centering
  \includegraphics[width=0.78\columnwidth]{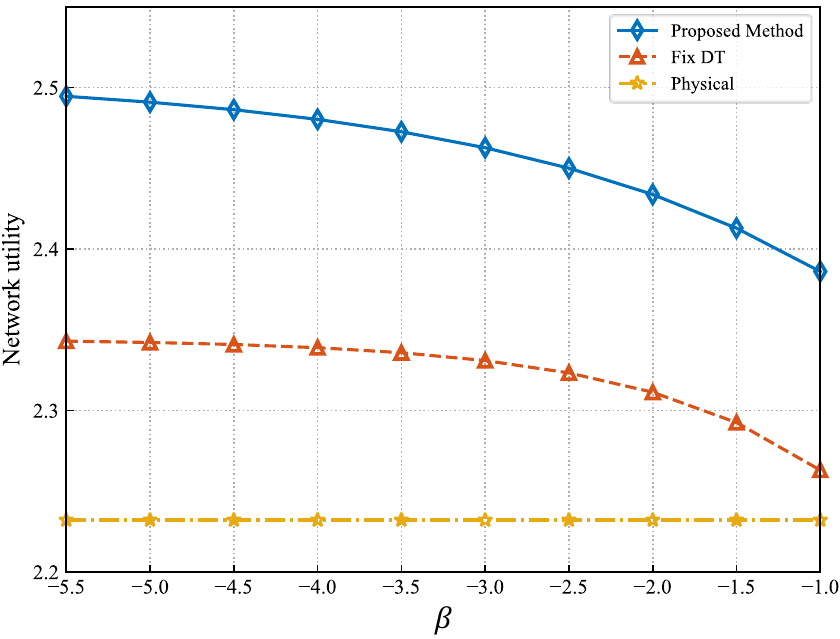}
   \vspace{-5pt}
  \centering \caption{Network utility v.s. $\beta$.}
  \label{fig5}
   \vspace{-5pt}
\end{figure}

Fig. \ref{fig5} shows the network utility with respect to the weight on the cost, i.e., $\beta$. Since cost is negatively correlated with noise, $\beta$ takes a value that is always negative. It can be observed that as the absolute value of $\beta$ increases, the advantage of the DT method becomes more obvious. This is due to the fact that a larger absolute value of $\beta$ means that the cost decreases faster as the noise increases, while the cost of the physical UAV does not change, so the cost advantage of the DT method is more prominent. Similarly, the choice of DT by RL allows it to achieve a network utility better than other methods.

\section{Conclusion}
In this paper, to reduce the DRL training cost in multi-UAV networks caused by buying UAVs and the flying energy cost while guaranteeing the DRL training performance of optimizing the UAV trajectory, an imperfect DT-assisted low-cost reinforcement method has been proposed, where there are deviations between physical and digital space in DT. The joint problem of reducing training costs and increasing the average transmission rate of UAV networks in time is solved by the mixing optimization of scalar variables $K$ and $\delta$ and the time serial variable $\mathbf{v}$, where two cascade NNs are trained by the low-cost label-free unsupervised learning and reinforcement learning. Simulation results show the proposed method significantly reduces the DRL training cost of practical training, e.g., energy and hardware purchases, and guarantees a high network transmission rate by optimizing UAV trajectory. By applying the proposed method in the network, the training cost of the DRL-based optimization can be reduced even with imperfect DT, which implies that an efficient decision can be made with imperfect DTs in multi-UAV networks. For further research, we will explore how to prove the benefits of using imperfect DT for DRL training.

\section*{Acknowledge}
This work was supported by the National Key Research and Development Program of China (2020YFB1807700), and the National Natural Science Foundation of China (NSFC) under Grant No. 62071356.

\bibliography{ref}
\bibliographystyle{IEEEtran}

\ifCLASSOPTIONcaptionsoff
  \newpage
\fi

\end{document}